\documentclass{article}
\usepackage[utf8]{inputenc}
\usepackage{authblk}
\usepackage{setspace}
\usepackage[margin=1.25in]{geometry}
\usepackage{graphicx}
\graphicspath{ {./figures/} }
\usepackage{subcaption}
\usepackage{amsmath}
\usepackage{lineno}
\usepackage{booktabs}
\usepackage{algorithm}
\usepackage{algpseudocode}
\usepackage{xcolor}
\usepackage{pdfpages}
\usepackage{multirow}
\usepackage{array}
\usepackage{makecell} 
\usepackage{booktabs} 
\usepackage{url} 
\usepackage[table]{xcolor}

\usepackage[backend=biber,
style=numeric]{biblatex}
\title{ViViD-5K: Vineyard vision dataset for field-based berry detection and segmentation and grape cluster closure estimation
}

\author[1$\dag$]{Xiangzhi Tong}
\author[2$\dag$]{Chengrui Zhang}
\author[3]{Mac Flaherty}
\author[3]{Andre Matteo Garcia}
\author[3]{Dominic Gorman}
\author[1]{Jonathan Jaramillo}
\author[3]{Justine E. Vanden Heuvel}
\author[4*]{Yu Jiang}

\affil[1]{School of Electrical and Computer Engineering, Cornell University, NY, USA.}
\affil[2]{Department of Electrical Engineering and Electronics, University of Liverpool, Liverpool, UK.}
\affil[3]{Horticulture Section, School of Integrative Plant Science, Cornell University, Ithaca, NY, USA.}
\affil[4]{Horticulture Section, School of Integrative Plant Science, Cornell AgriTech, Cornell University, Geneva, NY, USA.}

\affil[*]{Correspondence to: yujiang@cornell.edu}
\affil[$\dag$]{These authors contributed equally to this work.}

\date{}

\begin{document}
\maketitle

\begin{abstract}
    
Cluster closure, defined as the progressive filling of gaps between the berries in a grape bunch, is a key trait in vineyard management, impacting disease risk. However, traditional visual scoring methods are labor-intensive, subjective, and lack temporal resolution. Existing datasets rarely support fine-grained berry-level analysis, limiting the development of robust deep learning models. In this work, we present \textbf{ViViD-5k}, a large-scale in-field \textbf{Vi}neyard \textbf{Vi}sion \textbf{D}ataset containing 5,000 images with dense annotations, including over 648,000 berry centroids and cluster segmentation masks spanning 13 grape varieties. Building on this dataset, we introduce GrapeSAM, a two-stage visual pipeline that combines point-based berry localization with prompt-based segmentation using Segment Anything, followed by transformer-based cluster segmentation. The pipeline enables automated, in-field estimation of cluster closure with minimal supervision. Quantitative results demonstrate strong segmentation and counting accuracy across diverse conditions, while visualizations confirm robustness on both in-domain and out-of-domain samples. This work provides a scalable and objective alternative to manual compactness scoring and supports high-throughput grape phenotyping with enhanced spatial detail.

\end{abstract}


\section{Introduction}

Grape cluster closure (CC) refers to the phenological stage at which berries grow large enough to touch each other, filling in the gaps within a cluster. This phenological stage was reported to correspond with the modified stage 32 of Eichhorn Lorenz (E-L) \cite{coombe1995adoption}, although CC is better thought of as a process rather than a specific stage \cite{trivedi2023preliminary}. Grape cluster closure is significant in vineyard management because, once clusters fully close, it becomes harder for sprays to penetrate, and the microclimate inside the bunch changes, impacting disease development. This phenological stage, however, takes place over many weeks as the grape cluster changes from a few berries touching each other to a fully closed structure where every berry is in full contact with its neighboring berries. This evokes the need for a metric by which cluster closure can be quantified and monitored throughout its phenological development.

Grape cluster compactness, a widely used trait related to, yet distinct from, CC, is defined as the density of berries in a grape bunch compared to the length of the rachis \cite{sepahi2016estimating, christodoulou1967response} and influences berry integrity and cluster disease susceptibility. Highly compact clusters can lead to increased risk of fungal infections due to poor air circulation and microcracks in the berry cuticle membranes as they approach maturity, ultimately affecting grape and wine quality \cite{hed2009relationship, vail1991grape, tello2014evaluation}. Furthermore, grape cluster compactness shows impact on water displacement \cite{sepahi2016estimating}.

The International Organization of Vine and Wine (OIV) defines a compactness descriptor (OIV code 204) that classifies clusters into five classes, from very loose to very compact, based on morphological cues such as berry spacing and pedicel visibility. While widely used in grape research \cite{palliotti2011early, tardaguila2012mechanical}, this visual rating method is laborious and subjective. Additionally, this visual rating only reflects the general tightness of grape cluster but lacks the consideration of either rachis length for grape cluster compactness or the dynamic process of cluster closure over time. 

Recent studies have focused on the development of imaging-based methods to improve the accuracy, throughput, or both for measuring grape cluster compactness and closure. Palacios \cite{palacios2019non} investigated the use of color and visual features extracted by speeded-up robust features (SURF) to obtain masks of grape clusters, berries, and rachis to calculate the cluster compactness. Experimental results showed good correlation ($R^2=0.68$) between image-derived compactness and OIV ratings. Underhill \cite{underhill2020image_sys, underhill2020image} improved the measurement accuracy ($R^2=0.9$) by enhancing the illumination consistency and therefore color feature robustness through a customized imaging system. Kim \cite{kim2023shine} used conventional image processing to obtain the mask of a grape cluster and a Mask RCNN model to get the masks of individual berries within the cluster. Subsequently, density of grape cluster (DGC) was calculated as the ratio of pixels of gap space between berries within a cluster to that of the entire cluster. Cluster and berry segmentation was highly accurate ($mAP=0.96$). The individual berry masks achieved an average measurement error of less than 5\% for berry diameter estimation, and the DGC metric showed quantitative differences among grpae clusters. These achievements were primarily due to the combined advantage of a illumination-controlled imaging system in the lab and advanced deep learning models. This imaging processing pipeline requires controlled illumination for image acquisition, limiting its potential for applications in viticulture. 

It is noteworthy that DGC does not precisely correspond to either cluster compactness or cluster closure as defined in biological terms, as it neither normalizes berry density by rachis length to represent compactness nor captures the progressive changes of cluster closure. It is also unrelated to the classical definition of density: the ratio of mass to volume. To address potential confusion and clarify terminology, we propose a new term: \textbf{visual cluster closure (VCC)}, which better characterizes the metric of the ratio of pixels of gap space between berries within a cluster to that of the entire cluster in an image. In lieu of manually measuring grape cluster compactness or the OIV compactness descriptor, VCC offers a simple and effective way to characterize CC progress throughout the growing season \cite{kim2023shine, trivedi2023preliminary}.

Trivedi \cite{trivedi2023preliminary} extended the capacity of VCC measurement to field-based applications and first explored the use of asymptotic regression to characterize the CC progress. Images were acquired using smartphones with a handy whiteboard as background to generate high contrast between grape clusters and the background. Their imaging analysis pipeline leveraged the pyramid scene parsing (PSP) network to accurately segment grape clusters ($mIoU>0.9$) and the Otsu's thresholding to identify gap space between berries within a cluster ($<2\% error$). The VCC values over time showed an asymptotic trend: a higher rate of progression observed in the first three weeks followed by a gradual approach towards an asymptote. This method provides a continuous scale to describe CC progression throughout a growing season and tightly aligns with the CC biological definition, despite of the limitation of using a whiteboard as background for operation.

Through these research efforts, object detection and image segmentation are widely recognized as the key to accurate and automated measurement of grape cluster compactness and closure, including detection of grape cluster and berry, semantic segmentation between grape cluster, non-grape cluster, and gap space, and instance segmentation of individual grape clusters and berries. Pilot studies focused on classical machine learning or statistical models based on image features, hough transforms, or structured light for visual detection of berries \cite{aquino2017new, nuske2011yield}, owing to improved model performance \cite{chen2023instance, coviello2020gbcnet, du2023instance, gonzalez2025comparison}, recent research has shifted to the use of deep learning models to calculate variables related to cluster compactness and closure such as berry counting \cite{zabawa2020counting, aquino2017new, coviello2020gbcnet} and rachis length estimation \cite{huang2013procedural, xin20223d}. While these models are able to provide accurate detection and segmentation results for images with complex background (e.g., in the vineyard), a significant amount of annotated datasets is required to train these models, validate model performance, and improve model generalizability. To the best of our knowledge, Embrapa WGISD~\cite{santos2020grape}, the largest publicly available dataset with berry-level annotation, contains only 300 images, hindering the development of robust and versatile deep learning models for viticulture research and management.


The overarching goal of this study was to facilitate the development and evaluation of deep learning-based grape cluster analysis by addressing the data scarcity issue and providing an open-source analysis software. Specific objectives were to i) prepare and curate a dataset of 5,000 field-collected grape cluster images with annotations including cluster bounding boxes, masks, and berry point labels, ii) train and evaluate a baseline segmentation model for detecting grape clusters and berries, and iii) develop an automated processing pipeline that integrates the baseline model to enable grape cluster closure analysis.


\section{Vineyard Vision Dataset 5K (ViViD-5K)}
Despite the increasing interest in computer vision for viticulture applications, the availability of well-annotated, large-scale image datasets remains extremely limited. Many existing datasets, including Grape CS-ML [2018]~\cite{seng2018computer}, GrapesNet [2023]~\cite{BARBOLE2023109100}, and wGrapeUNIPD-DL [2022]~\cite{SOZZI2022108466}, provide only raw images with minimal or no annotation, limiting their uses for training and evaluating deep learning models especially via supervised fashion (Table~\ref{tab:vineyard-summary}). Few datasets are manually annotated, but their annotations are often constrained to a single task (e.g., berry counting or cluster detection) and barely provide comprehensive annotations that can be used for multiple computer vision tasks. For example, the segmentation of wine berries dataset~\cite{zabawa2021segmentation} offers only berry point labels for berry detection and counting, while CERTH ~\cite{agronomy13081995} provides cluster-level masks for grape cluster segmentation. Besides the comprehensiveness of annotation types, the existing datasets are usually collected for a limited number of grape varieties and cannot cover sufficient variability in grape appearance, shape, and size caused by natural variety differences. This bottleneck makes it difficult to build models that generalize well across grape varieties or grapes from diverse geographic regions. This data scarcity issue hinders model development and evaluation for improving performance of a given computer vision task and/or for building joint models that can solve multiple computer vision tasks simultaneously.

\begin{table}[htbp]
\caption{Vineyard Datasets}
\label{tab:vineyard-summary}
\centering
\resizebox{\textwidth}{!}{%
\begin{tabular}{@{}>{\raggedright\arraybackslash}m{4.0cm} 
                  >{\centering\arraybackslash}m{1.2cm} 
                  >{\centering\arraybackslash}m{1.2cm} 
                  >{\centering\arraybackslash}m{2.0cm} 
                  >{\centering\arraybackslash}m{2.5cm} 
                  >{\centering\arraybackslash}m{2.5cm} 
                  >{\centering\arraybackslash}m{2.7cm} 
                  >{\centering\arraybackslash}m{1.8cm}@{}}
\toprule
\textbf{Name} & \textbf{Year} & \textbf{Num\#} & \makecell{\textbf{Berry} \\ \textbf{Points}} & \makecell{\textbf{Cluster} \\ \textbf{Bounding Box}} & \makecell{\textbf{Cluster} \\ \textbf{Segmentation}} & \makecell{\textbf{Scenario} \\ \textbf{\& View}} & \makecell{\textbf{Grape} \\ \textbf{Varieties}} \\ 
\midrule

Grape CS-ML(1–4) \cite{seng2018computer} & 2018 & 2,016 & \makecell{-} & \makecell{-} & \makecell{-} & Outdoor; Close & 15 \\

Grape CS-ML(5) \cite{seng2018computer} & 2018 & 62 & \makecell{-} & \makecell{-} & \makecell{-} & Outdoor; Close & 1 \\

Embrapa WGISD  \cite{santos2020grape} & 2019 & 300 & 7,349 & 4,431 & 2,020 & Outdoor; Distant & 5 \\

AI4EU \footnotemark[1] \cite{morros2021ai4agriculture}  & 2021 & 250 & \makecell{-} & 5,076 & \makecell{-} & Outdoor; Distant & 1 \\

\makecell[l]{Segmentation of\\Wine Berries} \cite{zabawa2021segmentation} & 2021 & 42 & 21,195 & \makecell{-} & \makecell{-} & Outdoor; Distant & 3 \\

wGrapeUNIPD-DL \footnotemark[1] \cite{SOZZI2022108466}  & 2022 & 271 & \makecell{-} & 2,197 & \makecell{-} & Outdoor; Distant & 4 \\

\makecell[l]{Grapevine Bunch\\ Detection} \footnotemark[1] \cite{isabel_pinheiro_2023_7717055} & 2023 & 910 & \makecell{-} & 1,066 & \makecell{-} & Outdoor; Close & 6 \\

GrapesNet\cite{BARBOLE2023109100} & 2023 & 2,129 & \makecell{-} & \makecell{-} & \makecell{-} & Outdoor; Distant & 1 \\

CERTH \footnotemark[1] \cite{agronomy13081995}  & 2023 & 2,502 & \makecell{-} & 9,832 & 9,832 & Outdoor; Close & 1 \\

GrapeSet \footnotemark[2] \cite{jimaging11020034} & 2025 & 2,160 & 997,202 & \makecell{-} & 891 & Indoor; Close & 3 \\
\textbf{ViViD-5k (Ours)} & \textbf{2025} & \textbf{5,000} & \textbf{648,710} & \textbf{18,575} & \textbf{18,575} & \textbf{Outdoor; Both} & \textbf{13} \\ 
\bottomrule
\end{tabular}
}
\end{table}

\footnotetext[1]{Further annotated by the authors, constituting a subset of ViViD-5k.}
\footnotetext[2]{Only berry counting numbers, no labeling.}

To overcome the data scarcity challenge, we introduce vineyard vision dataset 5K (ViViD-5K), an image dataset of annotated vineyard images for grape cluster analysis. ViViD-5K comprises 5,000 grapevine images captured in real-world vineyard conditions and provides bounding boxes and masks for grape cluster detection and segmentation and point labels for individual berry detection and possibly instance segmentation. To leverage existing data sources, a total of 4,035 out of 5,000 images were selected from previous studies, where existing annotations were refined and/or expanded to include cluster masks or berry point labels. The remaining 965 images were collected and annotated by the authors to enrich the diverse representation of grapevine appearances of various grape varieties across different conditions. Images from the authors were acquired using smartphone cameras at research vineyards of Cornell University across Ithaca, Lansing, and Geneva in New York. Image acquisition commenced immediately after berry set under natural lighting conditions and was conducted regularly (roughly weekly) depending on the cultivar. The target grape cluster was consistently centered in each image. The whole dataset was labeled with cluster masks and berry point labels by four human annotators for 3284 hours using the CVAT platform$~^3$\footnotetext[3]{CVAT: Computer Vision Annotation Tool. Available at: \url{https://www.cvat.ai/}}. Combining all these datasets resulted in 50\% of the images from the CERTH dataset~\cite{agronomy13081995}, 18.2\% from Grapevine Bunch Detection~\cite{isabel_pinheiro_2023_7717055}, 7.5\% from wGrapeUNIPD-DL~\cite{SOZZI2022108466}, 4.9\% from AI4EU~\cite{morros2021ai4agriculture}, and 19.4\% of images from the authors (Figure~\ref{fig:vivid-composition}). 

\begin{figure}[H]
  \centering
  \includegraphics[width=0.7\textwidth]{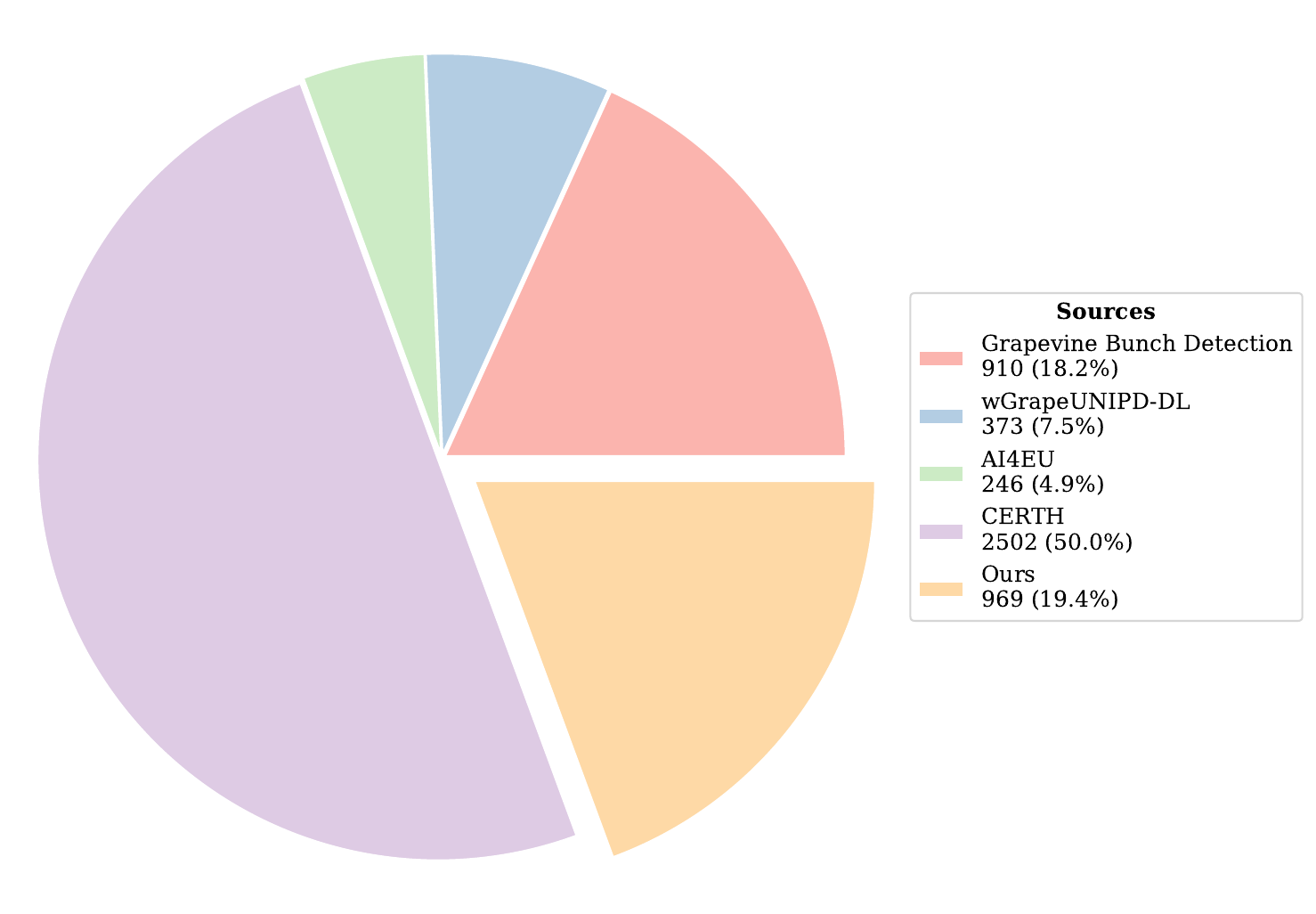}
   \caption{Image source distribution of ViViD-5K. It is noteworthy that although majority of raw images are from other sources, their annotations are compiled by the authors to form ViViD-5K. Additionally, images provided by the authors include diverse grape varieties that are economically important to the grape and wine industry worldwide.}
    \label{fig:vivid-composition}
\end{figure}

Based on the data acquisition distance, the ViViD-5K images can be categorized as “close-up”, composed of predominantly a single grape cluster in an image, or “distant-view”, composed of multiple grape clusters in an image. The majority (87.62\%) of dataset is close-up images, and the remaining 12.38\% images are distant-view including all from the AI4EU~\cite{morros2021ai4agriculture} and wGrapeUNIPD-DL datasets~\cite{SOZZI2022108466} and 514 images of immature grape clusters from the authors. 

\begin{figure}[H]
  \centering
  \includegraphics[width=0.9\textwidth, height=2.3in]{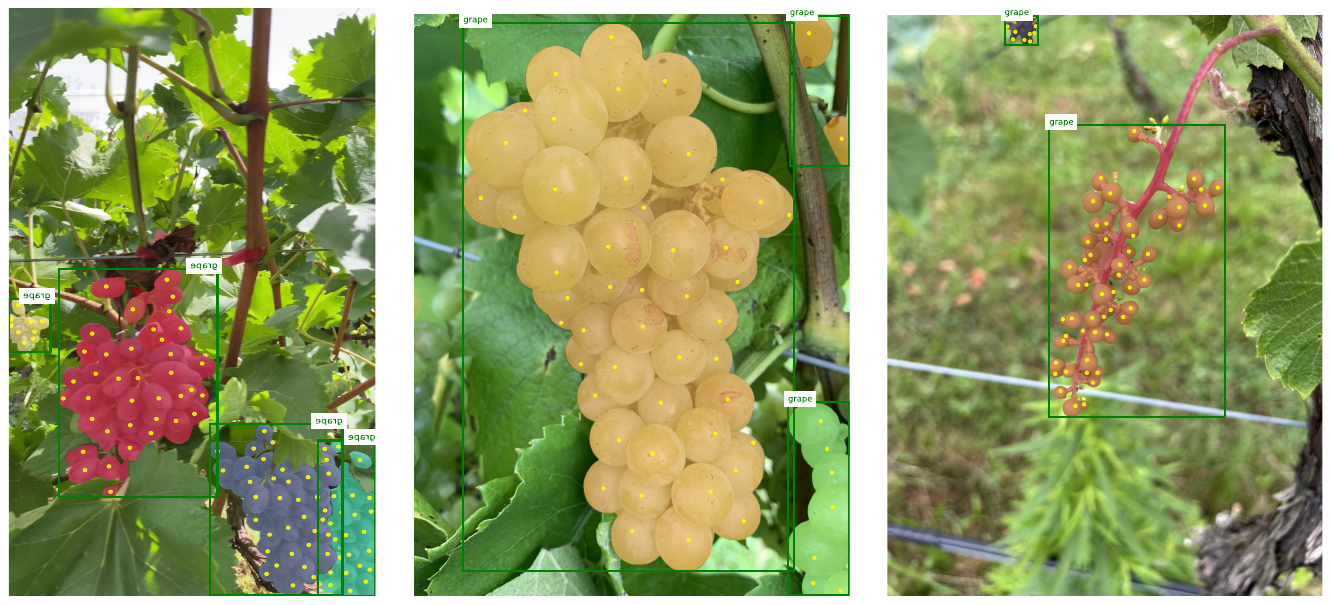}
   \caption[caption]{Visualization \footnotemark  of example images containing cluster bounding box and mask and berry point labels in ViViD-5K.}
    \label{fig:rep_annotation}
\end{figure}

All berry point annotations in our dataset were manually labeled from scratch. In contrast, the cluster instance segmentation masks were partially adapted and extended from the CERTH dataset, with approximately 53\% of the mask annotations originating from CERTH and the remaining through an active learning approach. Additionally, all bounding boxes were automatically derived the corresponding segmentation masks. Example images with grape cluster and berry annotations are shown in Figure~\ref{fig:rep_annotation}. 


\footnotetext{Different colors here and in the coming images only represent different instances.}

The distributions of grape cluster instances and berry points per image exhibit long tail patterns (Figure \ref{fig:2-distribution}). The average number of grape clusters per image is 3.8 (median of 3.0) with many images having more than 7 and up to 35 clusters per image. Similarly, the average number of berry points per image is 129.7 (median of 91.0) with many images having more than 300 and up to 2250 berries per image. This long tail distribution pattern highlights the sparsity and imbalance inherent in the dataset, presenting potential challenges for instance-aware learning in dense scenes.


\begin{figure}[H]
  \centering
  \includegraphics[width=0.9\textwidth, height=2.3in]{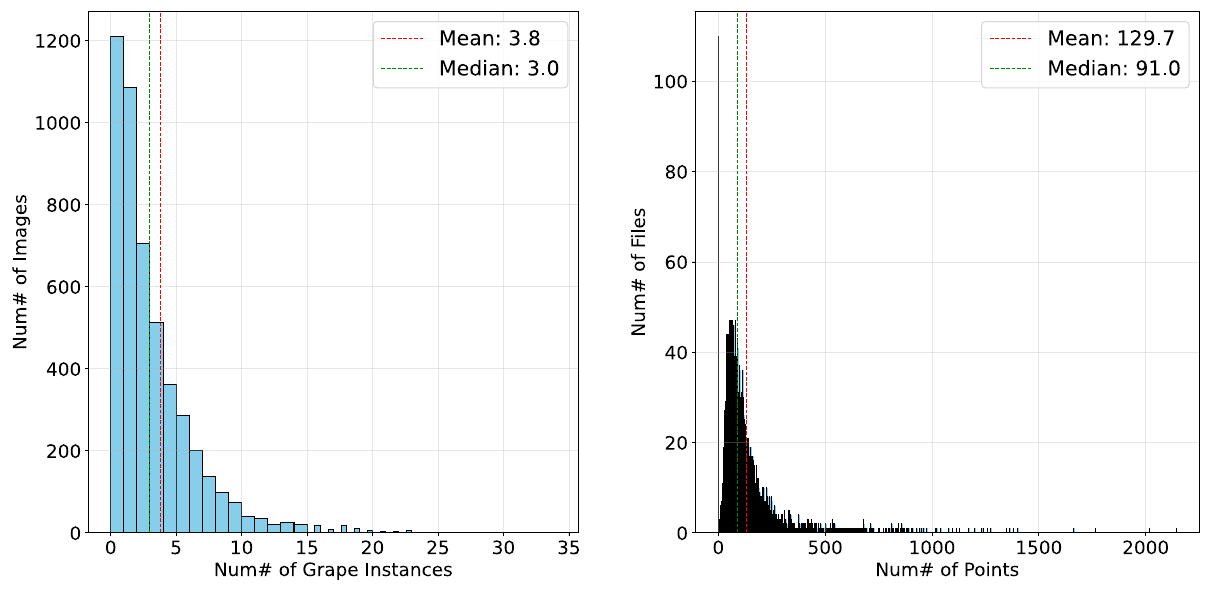}
   \caption{Grape and Berry Instance Frequency Count per File.}
    \label{fig:vivid-composition}
\end{figure}

All in all, ViViD-5K contains a diverse set of images collected using different systems (e.g., smartphones versus customized imaging systems), from mixed imaging distances (e.g., close-up versus distant-view), under both field and lab conditions with illumination variations, at different grapevine growth stages with varied levels of occlusion, and for various grape varieties (13 representative varieties). ViViD-5K is a valuable resource for developing and evaluating computer vision tasks in viticulture applications, including grape cluster and berry detection and segmentation for automated yield estimation and grape cluster analysis.

\section{Baseline approach for cluster and berry segmentation}
We leveraged the newly generated data in ViVid 5-k to develop a two-stage pipeline for computing in-situ VCC using a smartphone camera without the need for a whiteboard background. The following section outlines this pipeline and its application to measuring VCC.
\subsection{Preliminaries}
\subsubsection{Instance segmentation of grape cluster}
We adopted Mask2Former (Masked-attention Mask Transformer) \cite{cheng2022masked} for grape cluster instance segmentation due to its strong generalization capabilities and superior performance across diverse segmentation tasks. Mask2Former represents a state-of-the-art unified framework that seamlessly handles instance, semantic, and panoptic segmentation using a shared transformer-based architecture. For instance, segmentation in particular introduces a powerful query-based formulation, where each query is responsible for predicting a binary mask and its corresponding class label. These queries are iteratively refined across multiple transformer decoder layers, enabling the model to produce accurate instance masks even in cluttered or overlapping regions—common challenges in natural scenes such as grape clusters.

Its masked attention mechanism, which restricts attention to the most relevant spatial regions as defined by the predicted mask, not only improves computational efficiency but also enhances the model’s ability to focus on individual object instances, leading to sharper mask boundaries and reduced confusion in dense object layouts. To further capture spatial detail, the model incorporates a multi-scale feature pyramid and a high-resolution pixel decoder, which enable precise delineation of object boundaries. Compared to earlier models such as Mask R-CNN and MaskFormer, Mask2Former achieves state-of-the-art results on multiple benchmarks (e.g., COCO, ADE20K, Cityscapes), making it a compelling choice for high-precision instance segmentation in vineyard applications.

\subsubsection{Instance segmentation of berries}
Individual berry segmentation was performed using a two-stage methodology: i) Generate point detections corresponding to individual grape berry centroids (i.e., keypoints); and ii) Use these grape berry keypoints as prompts for segmentation using vision foundation models. We built our berry localization model based on a VGG19-based model \cite{Wan_2021_CVPR} with a custom dense heat map generation output head. This was inspired by crowd counting, a task that primarily focuses on estimating the centroid of human heads in dense scenes. In the second stage, the identified key points were passed into a prompt-based vision foundation model such as Segment Anything Model (SAM) \cite{kirillov2023segment} to derive the berry masks from point prompts. SAM is a powerful, prompt-driven segmentation framework designed for general applications. Pretrained on over one billion masks, it offers strong zero-shot generalization and supports flexible prompt modes such as bounding boxes, masks, and key points, making it ideal for generating dense grape berry masks from point prompts. It is composed of a vision transformer-based image encoder to extract rich hierarchical features from the raw image, while the prompt encoder embeds the prompts into the same feature embedding space, allowing another mask decoder module to fuse these embeddings to predict the segmentation masks conditioned by these prompts. In our baseline approach, each predicted berry centroid was passed into the SAM as the prompt to generate an instance-aware mask around the berry. Together, this two-stage approach enabled berry segmentation under field conditions with minimal supervision, making it well-suited for downstream tasks such as berry counting and cluster closure calculation.

\subsection{GrapeSAM}
\begin{figure}[H]
  \centering
  \includegraphics[width=1.0\textwidth]{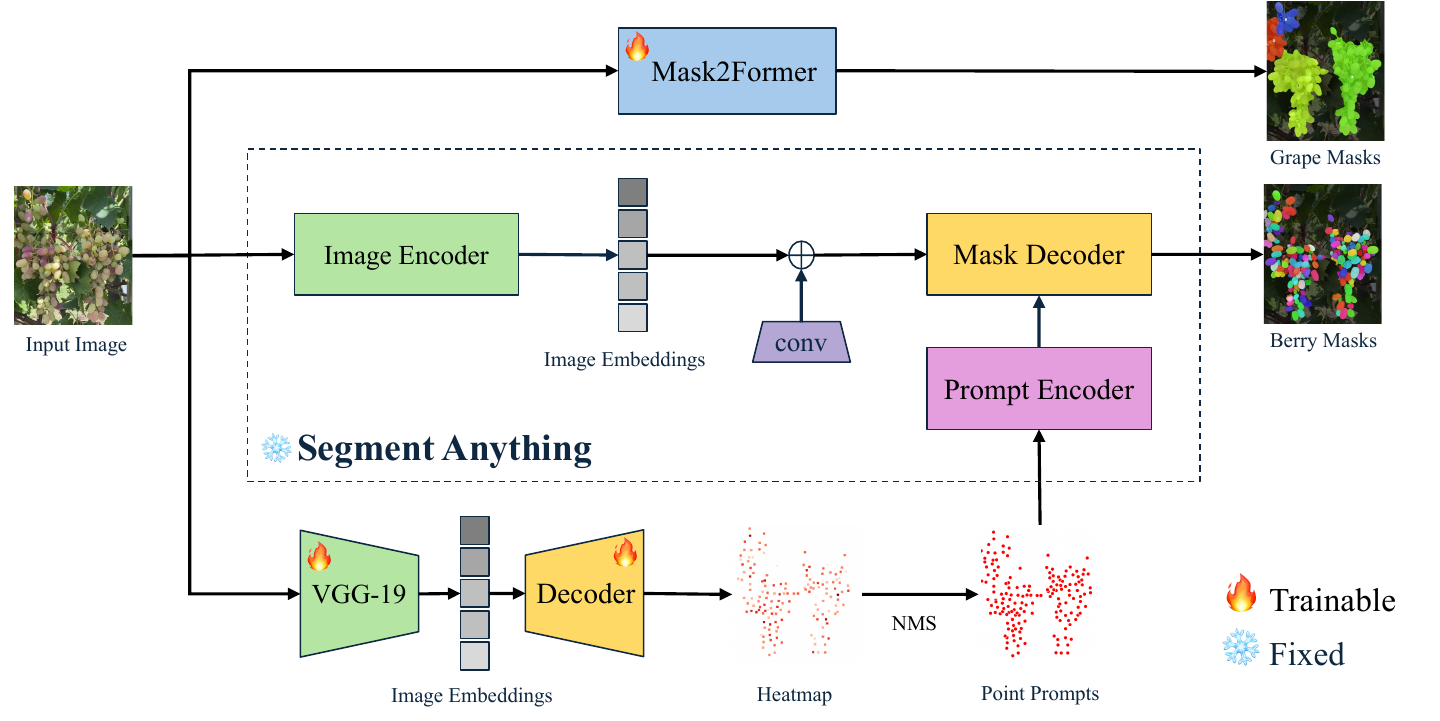}
  \caption{Diagram of GrapeSAM, the baseline approach for instance segmentation of grape clusters and individual berries in a given image. Model components with trainable parameters are indicated with the fire symbol, or otherwise the components are pretrained using general computer vision datasets. }
  \label{fig:grapesam}
\end{figure}

Our proposed baseline approach, GrapeSAM, comprises three parallel branches, each receiving the original input image simultaneously at the beginning of the pipeline (Figure \ref{fig:grapesam}). The Mask2Former branch is dedicated to performing whole-cluster instance segmentation of all grape clusters in the input image. In parallel, the point-based branch leverages a point decoder, which transforms image features extracted by its VGG19 backbone into a heatmap representation, revealing the localization probabilities of all individual berries. Non-Maximal Suppression (NMS) is used to extract the sparse spatial key points from the dense heatmap output. This is done by first upsampling the predicted heatmap $H \in [0,1]$, with spatial dimensions $\frac{H}{s} \times \frac{W}{s}$, by a factor of 8 using bilinear interpolation to match the resolution of the input image. A $3 \times 3$ max pooling operation is then applied across the spatial domain, which effectively suppresses redundant responses in local neighborhoods and retains only the local maxima as valid point candidates. Additionally, we also apply a confidence threshold $\tau$ (set to 0.05) on the suppressed heatmap to select candidate keypoints. All coordinates $(x, y)$ where the heatmap response exceeds $\tau$ are collected as candidate detections. Then these candidates are sorted in descending order according to their confidence scores, and only the top $K$ ($K=1024$ in this study) points are preserved. All detected points will be kept, if the total number of valid points is fewer than $K$. The resulting keypoints are further processed into a format compatible with SAM, where each point is wrapped in a singleton list to serve as a point prompt, and detailed berry-level masks are subsequently generated.

However, the resulting mask outputs frequently, albeit not invariably, exhibit redundant or duplicate instances. This issue typically manifests as a large, coarse mask overlapping and occluding multiple smaller, more accurate berry masks. To suppress such redundancy and preserve fine-grained instance details, we adopted an interquartile range (IQR)-based outlier removal mechanism as a dedicated post-processing step (Algorithm \ref{alg:iqr-filtering}). 

\begin{algorithm}
\caption{Mask Outlier Removal via IQR Filtering}
\label{alg:iqr-filtering}
\begin{algorithmic}[1]
\Require Mask set $\mathcal{M} = \{M_1, M_2, \ldots, M_n\}$
\Ensure Filtered mask set $\mathcal{M}'$ with outliers removed
\Function{FilterMasksByIQR}{$\mathcal{M}$}
    \State $a_i \gets \sum_{p \in M_i} \mathbf{1}(p), \forall i \in [1,n]$ \Comment{Mask areas}
    \State $\ell_i \gets \log(a_i + \epsilon), \forall i \in [1,n]$ \Comment{Log transform}
    \State $\hat{\ell}_i \gets \frac{\ell_i - \min(\ell)}{\max(\ell) - \min(\ell)}, \forall i \in [1,n]$ \Comment{Normalize}
    
    \State $Q_1, Q_3 \gets \text{Percentile}(\hat{\ell}, 25), \text{Percentile}(\hat{\ell}, 75)$
    \State $\text{IQR} \gets Q_3 - Q_1$
    
    \State $\mathcal{V} \gets \{i \mid Q_1 - 1.5 \cdot \text{IQR} \leq \hat{\ell}_i \leq Q_3 + 1.5 \cdot \text{IQR}\}$ \Comment{Valid indices}
    
    \State \Return $\{M_i \mid i \in \mathcal{V}\}$
\EndFunction
\end{algorithmic}
\end{algorithm}

Initially, the algorithm calculates the area of each mask and applies a logarithmic transformation to mitigate skewness in the distribution. The transformed values are then normalized to the range [0,1]. Next, the first quartile ($Q_{1}$) and the third quartile ($Q_{3}$) of the area distribution are calculated. Using these values, the interquartile range is defined as $IQR=Q_{3} - Q_{1}$, while the lower and upper bounds are set as  $Q_{1} - 1.5 \times \text{IQR}$ and $Q_{3}+1.5 \times \text{IQR}$, respectively. Mask areas falling outside these bounds are identified as outliers and subsequently removed. As a result, a subset of masks with areas within the acceptable range is obtained, ensuring that only masks with representative sizes are retained. This method is particularly effective for eliminating masks that are significantly smaller or larger than typical instances, thereby improving the robustness of downstream processing tasks.

Lastly, once we have the instance masks from both the cluster level: $S_{\text{cluster}}$ and the berry level: $S_i$, the grape cluster closure can be computed using Equation~\ref{eq:cc} defined in \cite{trivedi2023preliminary}.

\begin{equation} \label{eq:cc}
    \text{Visual Cluster Closure} = \frac{\text{\#no of berry pixel}}{\text{\#no of grape cluster pixel}} \times 100 = \frac{\sum_{i=0}^{N} S_i}{S_{\text{cluster}}} \times 100, 
\end{equation}

\noindent
where $N$ denotes the total number of detected berries.

\subsection{Model training}
We used the ViViD-5k dataset published in this work to construct our application. Given the diverse sources of images in our dataset, ensuring a consistent feature distribution across all subsets during training is essential. Each subset was stratified into training, validation, and test sets using an 8:1:1 ratio to achieve this. The final dataset was constructed by merging these subsets, resulting in 4,000, 500, and 500 images for training, validation, and testing, respectively.

All experiments are conducted on a workstation equipped with a single NVIDIA RTX 4090 GPU and 64 GB of system RAM. Our task comprises two separate subtasks: berry localization and grape cluster instance segmentation. For the berry localization model, our training setup follows the procedure described in~\cite{Wan_2021_CVPR}, with minor modifications. Specifically, we employ a batch size of 64 and train the model for 1500 epochs. The total training time amounts to approximately 41.6 GPU hours. For the grape segmentation model, we train Mask2Former using the Adam optimizer with a learning rate of $1 \times 10^{-4}$ and a batch size of 6. No image preprocessing techniques, such as cropping or resizing, are applied during training, as Mask2Former natively supports variable-resolution input. The total training duration for this stage is approximately 48 GPU hours.

\subsection{Performance evaluation}

We utilize the mean absolute error (MAE) and root mean squared error (RMSE) as standard evaluation metrics to assess berry localization performance, formulated as:
\begin{equation} 
    \label{eq:mae} 
    \text{MAE} = \frac{1}{n} \sum_{i=1}^n \left| y_i - \hat{y}_i \right|,
\end{equation}

\begin{equation} 
\label{eq:rmse} 
    \text{RMSE} = \sqrt{ \frac{1}{n} \sum_{i=1}^n \left( y_i - \hat{y}_i \right)^2 },
\end{equation}

\noindent where $n$ denotes the total number of samples, $y_i$ represents the ground truth value for the $i$-th sample, and $\hat{y}_i$ denotes the corresponding predicted value. 

These two metrics are widely adopted in object counting and regression tasks due to their interpretability and effectiveness in quantifying numerical prediction errors. MAE evaluates the average magnitude of the differences between the predicted and ground truth values, providing an intuitive measure of overall prediction accuracy regardless of error direction. In contrast, RMSE incorporates a squared penalty on the errors, making it more sensitive to larger deviations and thus reflecting the variance in prediction performance.

Mean intersection over union (mIoU), a widely used metric, is also used for evaluating semantic segmentation. Normaly, a higher mIoU value indicates better segmentation accuracy. 
\begin{equation} \label{eq:miou}
    \text{mIoU} = \frac{1}{N} \sum_{i=1}^N \frac{TP_i}{TP_i + FP_i + FN_i}
\end{equation}
Here, TP, FP, FN respectively stand for True Positive, False Positive, and False Negative; and N represents the total number of classes

Average precision (AP) over different Intersections over Union (IoU) thresholds (i.e., from 0.50 to 0.95 in steps of 0.05) is used to evaluate the performance of instance segmentation. The AP score is derived from the precision-recall curve:
\begin{equation} \label{eq:ap}
    AP = \int_{0}^{1} P(R) \, dR,
\end{equation}

\noindent where $p(r)$ denotes the precision as a function of recall $r$, and the integral computes the area under the precision-recall curve. Commonly reported AP metrics include: \textbf{AP@50}, \textbf{AP@75}, and \textbf{mAP} (\textit{AP@[.50:.95]}), corresponding to AP at an IoU threshold of 0.50, 0.75 and the mean AP averaged over IoU thresholds from 0.50 to 0.95 in increments of 0.05.

\section{Evaluation of the baseline approach}
\subsection{Performance of cluster and berry localization and segmentation}
Our berry localization model achieved an \textbf{MAE} of 21.46 and an \textbf{RMSE} of 54.59, reflecting a reliable estimate of berry count in varying cluster densities. For grape cluster segmentation, the Mask2Former model obtained a \textbf{mAP} of 54.99, with \textbf{AP@50} and \textbf{AP@75} reaching 78.12 and 59.15, respectively. Performance by object size further shows an  \textbf{APs}(small) of 0.07, \textbf{APm}(medium) of 21.20, and \textbf{APl}(large) of 60.42, indicating robust segmentation for large clusters but reduced accuracy on smaller or more compact instances.

In addition, we present representative results from the test subset of the dataset in Figure \ref{fig:in-domain-inference}. The visualizations confirm that the model accurately segments individual berries and clusters, even under challenging conditions such as occlusions and high object density. These results reinforce the quantitative findings, highlighting the pipeline’s ability to generalize well to in-domain samples with complex structural variability.

\textbf{\begin{figure}[H]
  \centering
  \includegraphics[width=0.83\textwidth]{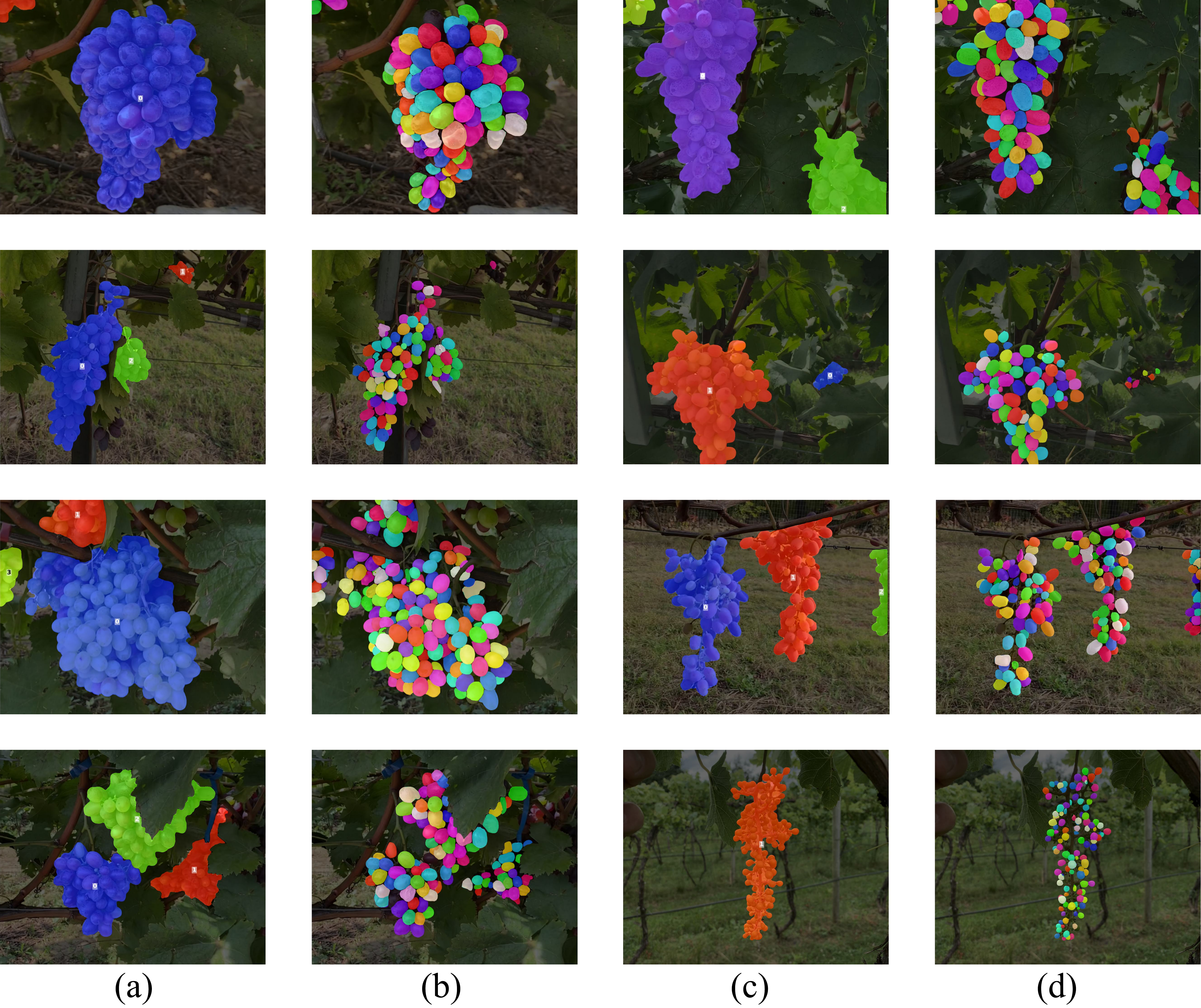}
  \caption{Representative inference results of in-domain images. \textbf{(a)} and \textbf{(c)} illustrate grape cluster segmentation results, whereas \textbf{(b)} and \textbf{(d)} show the corresponding berry-level segmentation results.}
  \label{fig:in-domain-inference}
\end{figure}}

\begin{figure}[H]
  \centering
  \includegraphics[width=0.7\textwidth]{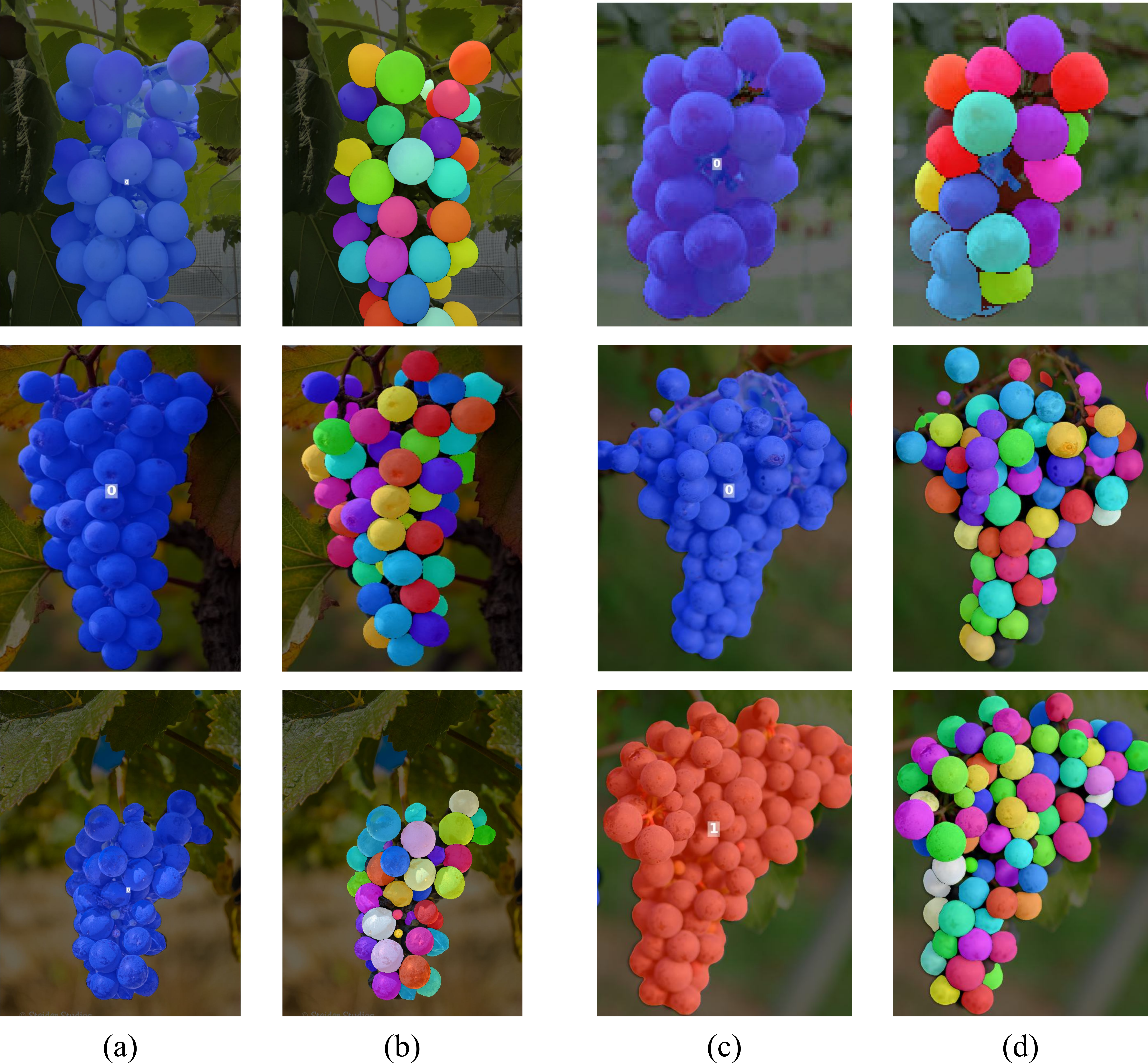}
  \caption{Representative inference results of out-of-domain images. \textbf{(a)} and \textbf{(c)} illustrate grape cluster segmentation results, whereas \textbf{(b)} and \textbf{(d)} show the corresponding berry-level segmentation results.}
  \label{fig:ood-inference}
\end{figure}

Given that our ultimate goal is to achieve strong generalization, it is vital to ensure the model performs robustly beyond the in-domain test field. Therefore, we further evaluated its performance on a set of previously unseen web-sourced images. These images were not included in the dataset and represent diverse real-world conditions. The corresponding segmentation results are visualized in Figure \ref{fig:ood-inference}. Despite variations in lighting, background clutter, and cluster morphology, the model remains capable of delineating both clusters and individual berries with consistent accuracy. In particular, the model demonstrates resilience to changes in berry coloration and occlusions. It performs robustly when dealing with conditions not explicitly represented in the training data.

\subsection{Ablation experiment for IQR-filtering algorithm}
\begin{figure}[H]
  \centering
  \includegraphics[width=0.8\textwidth]{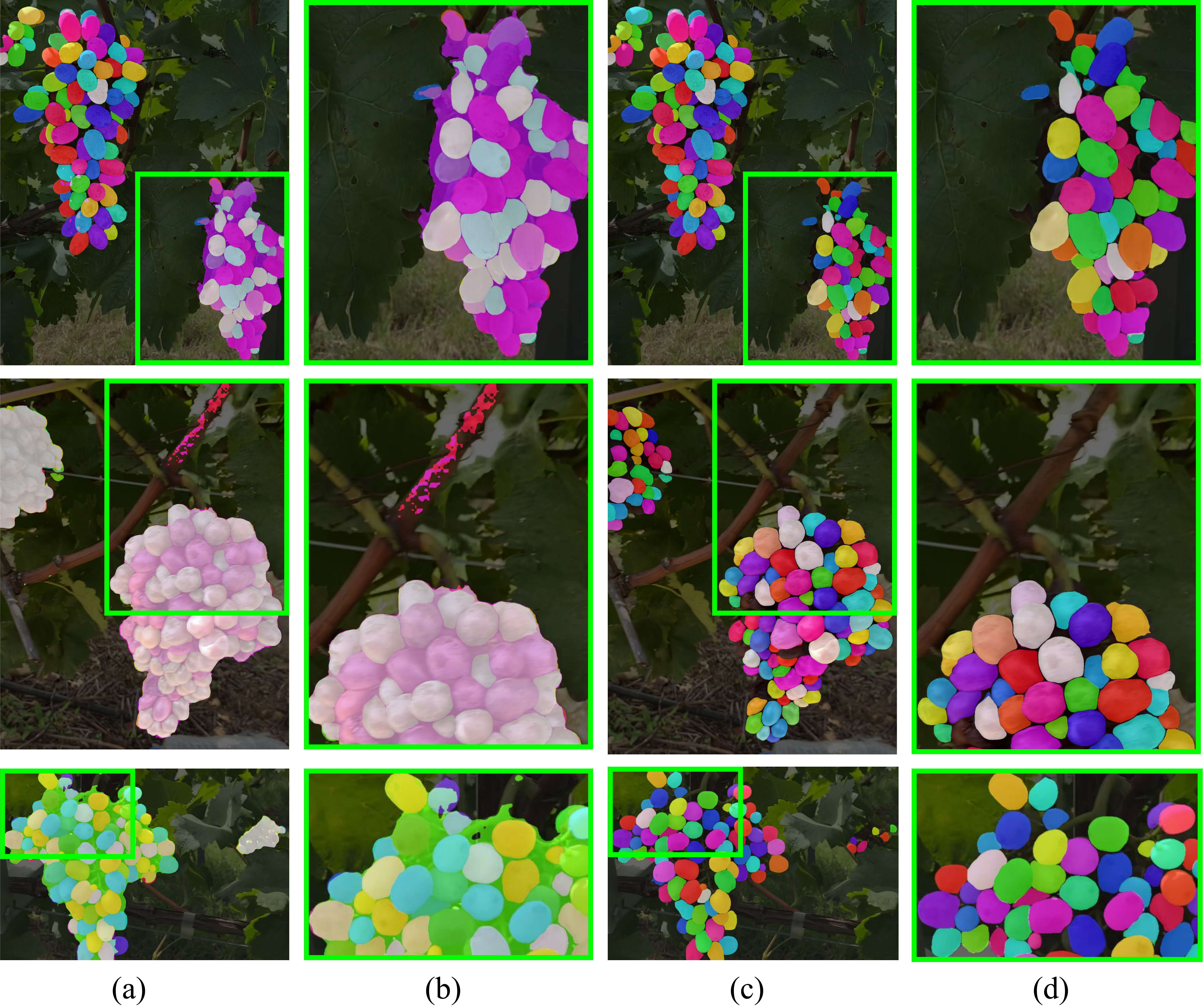}
  \caption{Qualitative results of the effectiveness of IQR-filtering.  Each row corresponds to a specific input image. The four columns are for \textbf{(a)} segmented masks of individual berries overlaid on grape cluster mask in the input image before IQR-filtering; \textbf{(b)} a zoomed-in view of the green bounding box of (a) in each image; \textbf{(c)} the masks of individual berries after IQR-filtering, and \textbf{(d)}a zoom-in view of the green bounding box of (c) in each image. Colors for individual berry masks are randomly assigned without semantics meaning.}
  \label{fig:iqr-vis}
\end{figure}

Although the developed baseline approach demonstrates satisfactory performance, it remains essential to validate the necessity and impact of the IQR filtering algorithm to ensure that it contributes meaningfully to the final segmentation quality. First, we visualize the representative samples before and after filtering in Figure \ref{fig:iqr-vis}. In the unfiltered predictions(\textbf{(a)} \& \textbf{(b)}), common artifacts that duplicate masks on adjacent berries, isolated false positives, and gaps left by suppressed tiny masks are clearly visible. Once the proposed IQR filter algorithm is applied (\textbf{(c)} \& \textbf{(d)}), these defects are largely removed. Spurious instances vanish, and duplicate masks are removed. Therefore, the resulting segmentations are markedly cleaner and topologically consistent. 

\begin{figure}[H]
  \centering
  \includegraphics[width=0.7\textwidth]{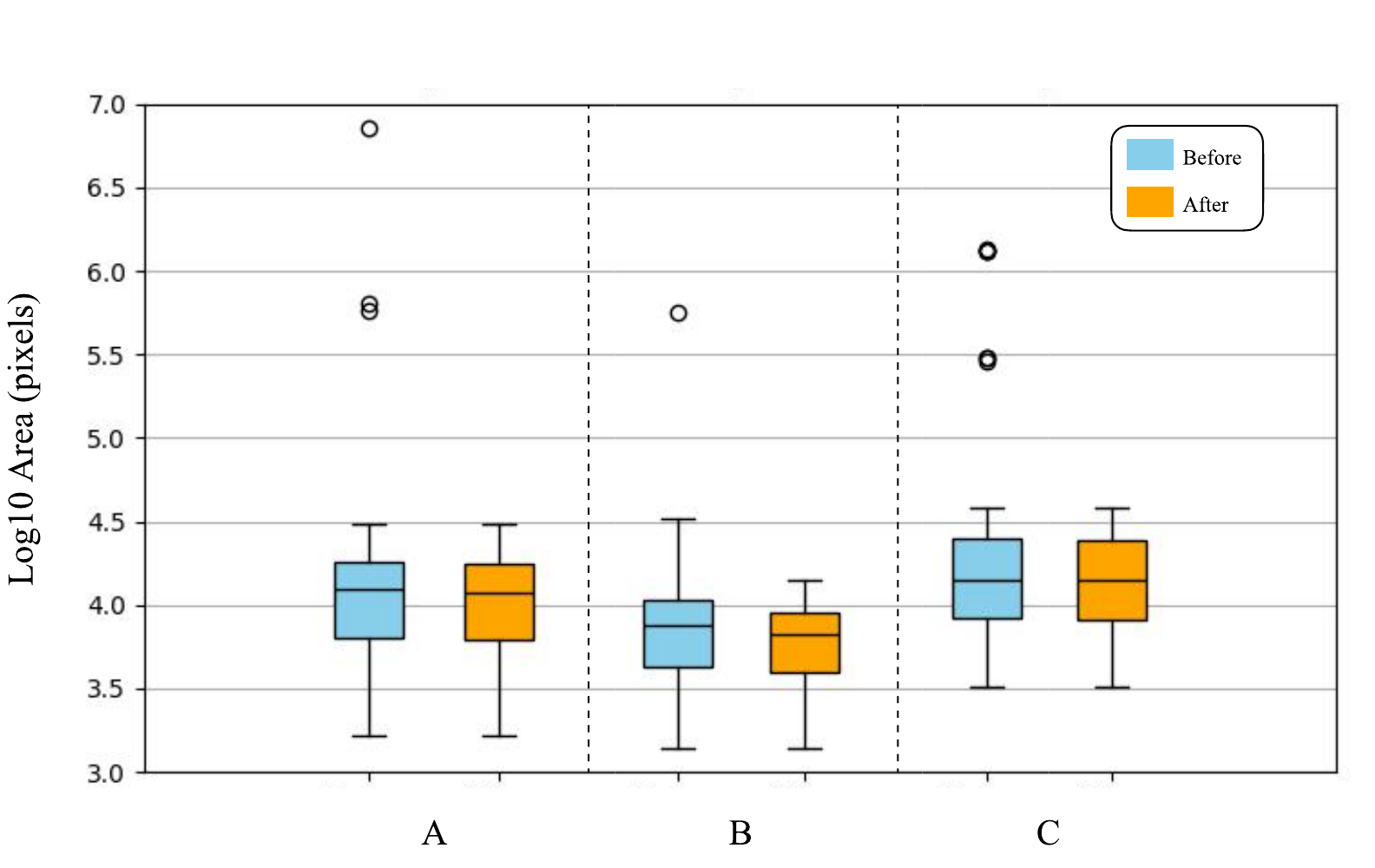}
  \caption{Boxplot of log-transformed instance area from individual berry masks segmented by the baseline approach with and without IQR-filtering. \textbf{A}, \textbf{B}, and \textbf{C} correspond to the 1st, 2nd, and 3rd row in Figure \ref{fig:iqr-vis}; Blue and orange represent pre- and post-filtering results. }
  \label{fig:filter-box-plot}
\end{figure}

Then, to quantitatively validate the effectiveness, Figure~\ref{fig:filter-box-plot} compares the distribution of connected component areas before and after the proposed area-based filter. Box-and-whisker plots show the base-10 logarithm of pixel area for the three evaluation samples visualized in Figure \ref{fig:iqr-vis}. Central horizontal lines indicate the median, the box spans the interquartile range (IQR), whiskers extend to $1.5 \times IQR$, and circles denote statistical outliers. Prior to filtering, the distribution exhibits a wide spread with extreme outlier values $\left( \log_{10} \text{ area} > 5.5, \; \text{i.e., } \text{area} > 3 \times 10^{5} \; \text{px} \right)$
, reflecting large spurious regions caused by merged masks.

\subsection{Case study for characterizing grape cluster closure}
To evaluate the usefulness of the ViViD-5K dataset and baseline approach, we analyzed an image dataset from a previous study~\cite{trivedi2023preliminary} to characterize grape cluster closure. This dataset was collected for a block of four contiguous panels (20 vines) of Pinot gris at the Cornell research vineyard in Lansing, NY, during the 2024 growing season. A total of 50 grape basal clusters were randomly selected from the 4 panels and imaged using smartphone cameras. This image collection was performed seven times from the berry set to the maturity stages. Although most images contained a single grape cluster centered in a close-up view, the images showed variations in cluster size due to natural growth differences, differences in imaging distance and angle, and changes in background. These factors reflect common challenges encountered in field applications. It should be noted that none of the grape cluster images were included in ViViD-5K.

\begin{figure}[H]
    \centering
    \begin{subfigure}{0.47\textwidth}
        \includegraphics[width=0.75\textwidth]{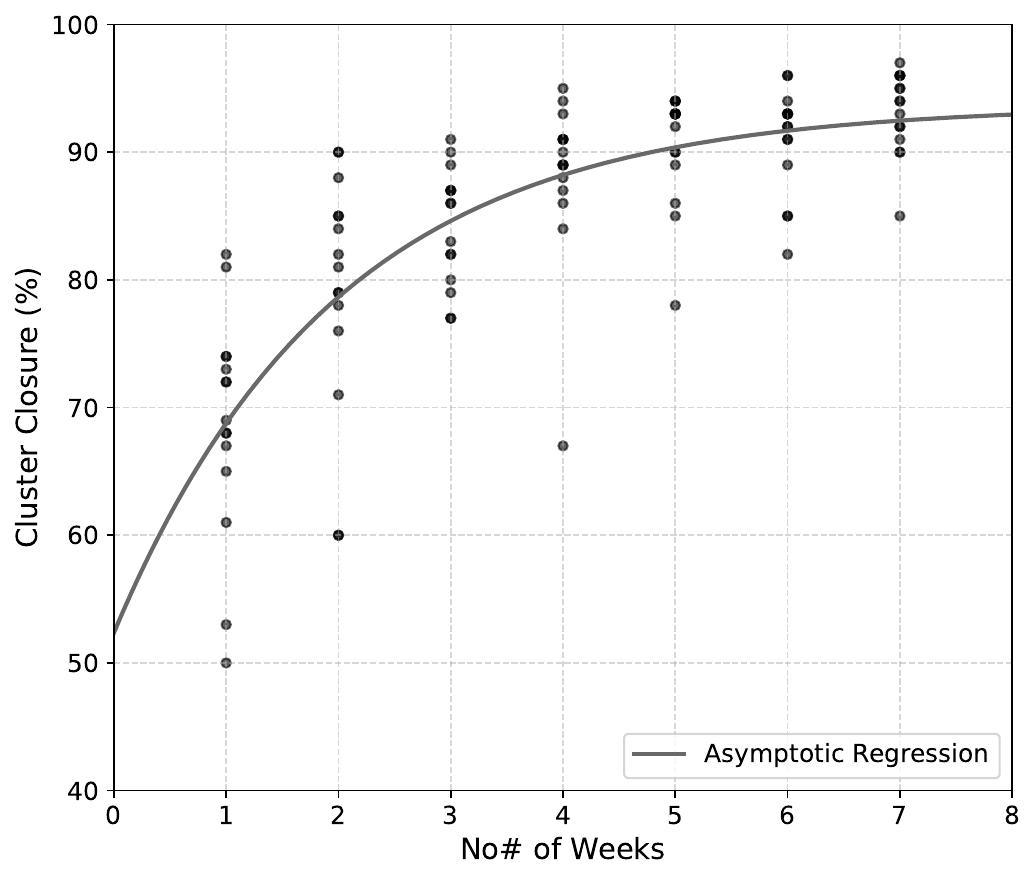}
        \caption{\label{fig:2a}}
    \end{subfigure}
    \begin{subfigure}{0.47\textwidth}
        \includegraphics[width=1.15\textwidth]{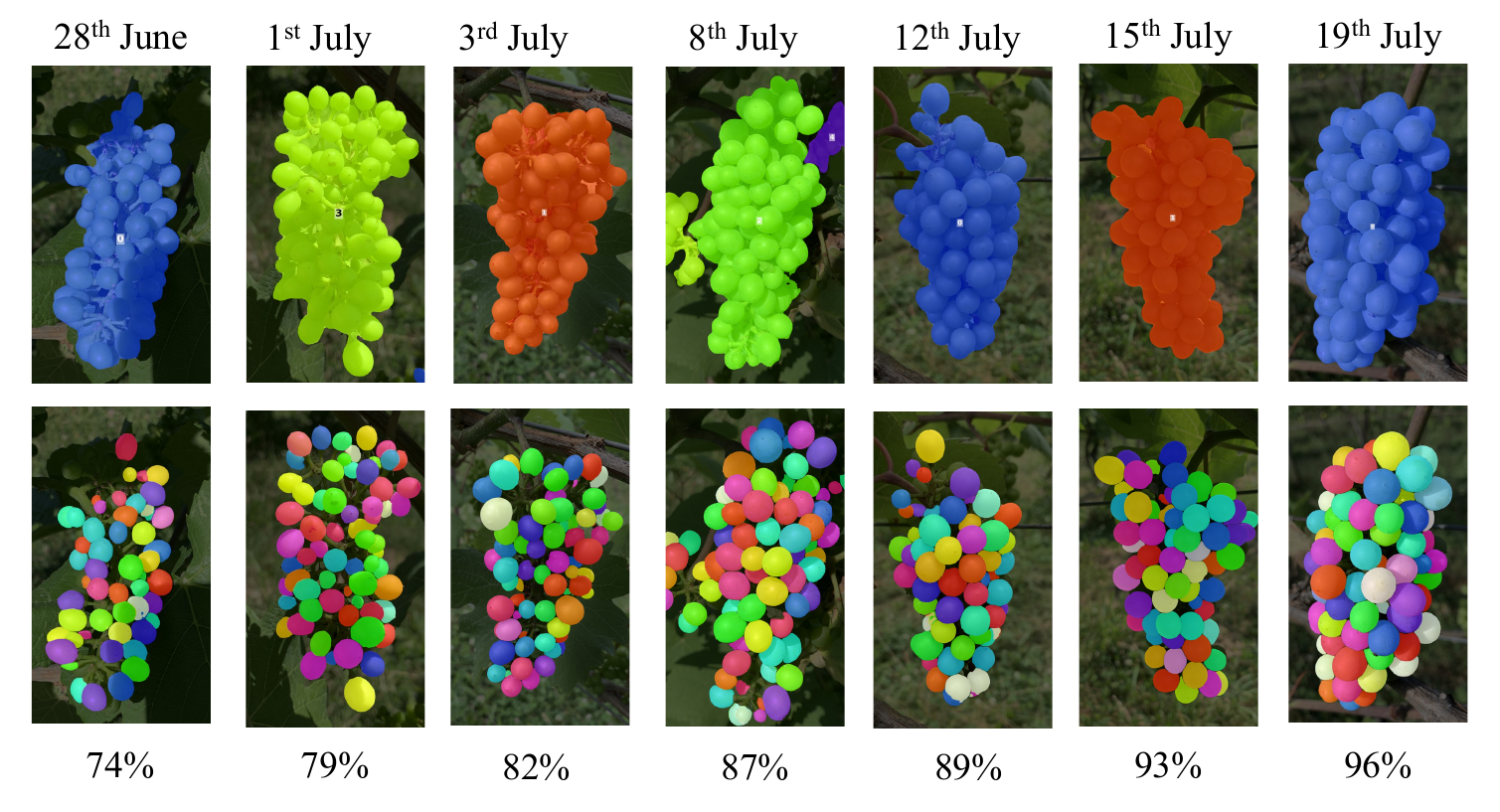}
        \caption{\label{fig:2b}}
    \end{subfigure}
    \caption{Basal Cluster Growth (\subref{fig:2a}) Cluster Closure Temporal Variation. (\subref{fig:2b}) Periodical Segmentation Samples.}
    \label{fig:2-distribution}
\end{figure}

We pretrained the GrapeSAM model using ViViD-5K and directly applied it to the aforementioned grape cluster images for zero-shot inference. The inference results included the masks of grape clusters and berries in individual images. For each given date, \%VCC was calculated using Equation~\ref{eq:cc}. Subsequently, an asymptotic regression model was used to fit the curve of \%iVCC over time for characterizing grape cluster closure, following the configuration defined in \cite{trivedi2023preliminary}. Compared with the original study~\cite{trivedi2023preliminary}, our baseline approach achieved comparable performance although no additional image labeling or model fine-tuning was used, demonstrating the value of having a large annotated dataset and pretrained model for future applications (Figure~\ref{fig:2-distribution}). Differences in model parameters between \cite{trivedi2023preliminary} and our study are likely due to weather patterns in the different study years (2020 in \cite{trivedi2023preliminary}, 2024 here) and locations of study blocks.

\begin{table}[h]
\centering
\caption{Asymptote regression model parameters comparison.}
\label{tab:asymptote_params}
\begin{tabular}{lcccc}
\toprule
\textbf{Cultivars} & \begin{tabular}[c]{@{}c@{}}Asymptote \end{tabular} & \begin{tabular}[c]{@{}c@{}}Intercept\end{tabular} & \begin{tabular}[c]{@{}c@{}}Rate of Increase\end{tabular} & \begin{tabular}[c]{@{}c@{}}Time of cluster closure \\ reaches Asymptote (weeks)\end{tabular} \\
\midrule
Manushi's \cite{trivedi2023preliminary}       & 94.58 & 42.24 & 1.29  & 3.46 \\
Ours & 93.67 & 52.36 & 0.506 & 4.31 \\
\bottomrule
\end{tabular}
\vspace{0.5em}
\end{table}


\section{Discussion}
Recent advances in semi-supervised and unsupervised learning have shown promising potential in addressing data scarcity challenges in computer vision. Nevertheless, the collection and curation of large annotated datasets remain critical and difficult tasks, not only for model development and training but also for reliable model evaluation. This challenge is especially pronounced in specialized domains such as plant phenotyping.

ViViD-5K was developed to address this gap. It contains 5,000 vineyard-collected images of 13 grape varieties, including instance bounding boxes and segmentation masks for 18,575 grape clusters, along with keypoint annotations for 648,000 individual berries. This dataset is expected to provide direct benefits to both the viticulture and plant phenomics research communities by supporting the advancement of deep learning models designed for grape cluster analysis. Moreover, the dataset holds value for benchmarking or inspiring the development of general computer vision models beyond this specific application.

A baseline method has been developed by integrating Mask2Former and GrapeSAM models, both pretrained on ViViD-5K, with a custom VGG-based point prompt network. This approach demonstrates the capability to segment grape clusters and individual berries simultaneously. The berry-level segmentation is achieved by leveraging keypoint labels as prompts in combination with vision foundation models such as SAM, which helps reduce the effort required for manual mask annotation at the berry level. While this method achieves satisfactory performance in downstream applications, it exhibits limitations when processing small and medium-sized objects. Specifically, there is a notable performance (i.e., AP) gap between small and large objects and between medium and large objects, with differences of 285 times and 2.85 times, respectively. These performance variations are more evident under varying imaging distances and angles. This observation further confirms the presence of a long tail distribution in the dataset, which is a widely recognized challenge in machine learning. Therefore, ViViD-5K represents a valuable benchmark for evaluating models and methods developed to address long tail distribution problems.

Through a case study on grape cluster closure analysis, the baseline method demonstrated strong generalization capabilities when applied to field-collected images. In a zero-shot inference setting, it successfully produced accurate segmentation masks of grape clusters and individual berries for calculating closure percentages over time. The results obtained were comparable to those generated by a model trained specifically on grape cluster data, highlighting the baseline method’s potential for immediate use in viticulture research and production management.

The case study also revealed two important issues that require attention in the plant phenomics community. First, the terms grape cluster compactness and grape cluster closure are sometimes used interchangeably, which may lead to confusion and miscommunication between viticulture and plant phenomics researchers. Although these traits are related, they are distinct. Cluster compactness describes berry density, which may suggest the degree of berry contact, but clusters with the same density can exhibit substantially different levels of berry contact. In contrast, cluster closure refers to the gaps between berries rather than overall density. Therefore, methods that directly estimate the number of berries without detection or segmentation would not be suitable for cluster closure analysis, which requires spatial information about the gaps between berries. Additionally, grape cluster closure is a dynamic trait that evolves over time, necessitating analysis methods that are robust to variations in imaging distance, viewing angle, and background conditions.

Despite the promising results achieved, both the ViViD 5K dataset and the baseline method have limitations that require continuous efforts. For example, in images where experimenters’ hands are visible, the segmentation model sometimes incorrectly includes parts of the hands as grape clusters, resulting in underestimation of closure measurements. Additionally, the berry localization model employs a lightweight architecture to accommodate hardware limitations, which reduces its effectiveness in scenes with high berry density or complex backgrounds. Model performance also decreases in images captured under different experimental conditions, such as indoor settings with uniform backgrounds. Although all annotations underwent quality control, challenges remain in accurately labeling instances with partial occlusion and high-density clustering. Furthermore, the dataset lacks balanced representation across all phenological stages, with immature grapes accounting for only approximately 10.4 percent of the dataset. This imbalance may affect model performance on early developmental stages. Finally, both the training process and annotation efforts remain resource intensive, indicating the need for further optimization.

\section{Conclusion}
A large image dataset, ViViD-5K, has been formed to share 5,000 grape cluster images of 13 representative grape variety with bounding box and mask annotations for grape clusters and keypoint labels for berries. A baseline approach, GrapeSAM, has been developed and evaluated for the characterization of grape cluster closure in zero-shot fashion, showing comparable performance to results obtained by models trained on the cluster closure dataset. This suggests both the usefulness of ViViD-5K and GrapeSAM for both viticulture applications and model development for grape cluster phenotyping in the future.


\section*{Acknowledgments}

\subsection*{General} 
We would like to thank Angela Paul for her support in data collection during this study, as well as for sharing her professional expertise in agricultural sciences. In addition, we thank Jiaqi Liu for his valuable contributions to the dataset annotation.

\subsection*{Author Contributions} 
X. Tong organized the dataset annotation, conceived the idea and built the code framework.
C. Zhang contributed equally to the software code. M. Flaherty and A. M. Garcia helped with the in-field data collection and data annotation. J. Jaramillo, J. E. V. Heuvel, and Y. Jiang advised this project and reviewed this paper.

\subsection*{Funding}
This project was jointly funded by a gift fund from Lorena Garmezy and Kevin Murphy to the Vanden Heuvel Lab and by the intramural research program of the U.S. Department of Agriculture, National Institute of Food and Agriculture, Multistate and Accession no. 7005734.

\subsection*{Conflicts of Interest}
The author(s) declare(s) that there is no conflict of interest regarding the publication of this article.

\subsection*{Data Availability}
The code will be released on GitHub at https://github.com/tong-xz/GrapeSAM. Additionally, the ViViD-5k dataset will be made public on huggingface: https://huggingface.co/datasets/XZhi/ViViD-5k.

\printbibliography

@article{sepahi2016estimating,
  title={Estimating duster compactness in Yaghouti grapes},
  author={Sepahi, A},
  journal={VITIS-Journal of Grapevine Research},
  volume={19},
  number={2},
  pages={81},
  year={2016}
}

@article{hed2009relationship,
  title={Relationship between cluster compactness and bunch rot in Vignoles grapes},
  author={Hed, Bryan and Ngugi, Henry K and Travis, James W},
  journal={Plant disease},
  volume={93},
  number={11},
  pages={1195--1201},
  year={2009},
  publisher={Am Phytopath Society}
}

@article{vail1991grape,
  title={Grape cluster architecture and the susceptibility of berries to Botrytis cinerea.},
  author={Vail, ME and Marois, JJ},
  journal={Phytopathology},
  volume={81},
  number={2},
  pages={188--191},
  year={1991}
}

@misc{coombe1995adoption,
  title={Adoption of a system for identifying grapevine growth stages. Aust J Grape Wine R},
  author={Coombe, BG},
  year={1995}
}

@article{trivedi2023preliminary,
  title={A Preliminary Method for Tracking In-Season Grapevine Cluster Closure Using Image Segmentation and Image Thresholding},
  author={Trivedi, Manushi and Zhou, Yuwei and Moon, Jonathan Hyun and Meyers, James and Jiang, Yu and Lu, Guoyu and Vanden Heuvel, Justine},
  journal={Australian Journal of Grape and Wine Research},
  volume={2023},
  number={1},
  pages={3923839},
  year={2023},
  publisher={Wiley Online Library}
}

@article{palliotti2011early,
  title={Early leaf removal to improve vineyard efficiency: gas exchange, source-to-sink balance, and reserve storage responses},
  author={Palliotti, Alberto and Gatti, Matteo and Poni, Stefano},
  journal={American Journal of Enology and Viticulture},
  volume={62},
  number={2},
  pages={219--228},
  year={2011},
  publisher={American Journal of Enology and Viticulture}
}

@article{tardaguila2012mechanical,
  title={Mechanical yield regulation in winegrapes: comparison of early defoliation and crop thinning},
  author={Tardaguila, Javier and Blanco, JA and Poni, Stefano and Diago, MP},
  journal={Australian Journal of Grape and Wine Research},
  volume={18},
  number={3},
  pages={344--352},
  year={2012},
  publisher={Wiley Online Library}
}

@article{christodoulou1967response,
  title={Response of Thompson Seedless grapes to prebloom thinning},
  author={Christodoulou, ARIS and Weaver, RJ and Pool, RM},
  journal={Vitis},
  volume={6},
  number={3},
  pages={303--308},
  year={1967}
}

@article{tello2014evaluation,
  title={Evaluation of indexes for the quantitative and objective estimation of grapevine bunch compactness},
  author={Tello, Javier and Ib{\'a}{\~n}ez Marcos, Javier},
  year={2014},
  publisher={Julius K{\"u}hn-Institut}
}

@article{zabawa2020counting,
  title={Counting of grapevine berries in images via semantic segmentation using convolutional neural networks},
  author={Zabawa, Laura and Kicherer, Anna and Klingbeil, Lasse and T{\"o}pfer, Reinhard and Kuhlmann, Heiner and Roscher, Ribana},
  journal={ISPRS Journal of Photogrammetry and Remote Sensing},
  volume={164},
  pages={73--83},
  year={2020},
  publisher={Elsevier}
}

@article{aquino2017new,
  title={A new methodology for estimating the grapevine-berry number per cluster using image analysis},
  author={Aquino, Arturo and Diago, Maria P and Mill{\'a}n, Borja and Tard{\'a}guila, Javier},
  journal={Biosystems engineering},
  volume={156},
  pages={80--95},
  year={2017},
  publisher={Elsevier}
}

@article{coviello2020gbcnet,
  title={GBCNet: In-field grape berries counting for yield estimation by dilated CNNs},
  author={Coviello, Luca and Cristoforetti, Marco and Jurman, Giuseppe and Furlanello, Cesare},
  journal={Applied Sciences},
  volume={10},
  number={14},
  pages={4870},
  year={2020},
  publisher={MDPI}
}

@article{xin20223d,
  title={A 3D grape bunch reconstruction pipeline based on constraint-based optimisation and restricted reconstruction grammar},
  author={Xin, Bolai and Whitty, Mark},
  journal={Computers and Electronics in Agriculture},
  volume={196},
  pages={106840},
  year={2022},
  publisher={Elsevier}
}

@article{huang2013procedural,
  title={Procedural grape bunch modeling},
  author={Huang, Chun-Yen and Jheng, Wan-Ting and Tai, Wen-Kai and Chang, Chin-Chen and Way, Der-Lor},
  journal={Computers \& graphics},
  volume={37},
  number={4},
  pages={225--237},
  year={2013},
  publisher={Elsevier}
}

@inproceedings{nuske2011yield,
  title={Yield estimation in vineyards by visual grape detection},
  author={Nuske, Stephen and Achar, Supreeth and Bates, Terry and Narasimhan, Srinivasa and Singh, Sanjiv},
  booktitle={2011 IEEE/RSJ International Conference on Intelligent Robots and Systems},
  pages={2352--2358},
  year={2011},
  organization={IEEE}
}

@article{chen2023instance,
  title={Instance segmentation and number counting of grape berry images based on deep learning},
  author={Chen, Yanmin and Li, Xiu and Jia, Mei and Li, Jiuliang and Hu, Tianyang and Luo, Jun},
  journal={Applied Sciences},
  volume={13},
  number={11},
  pages={6751},
  year={2023},
  publisher={MDPI}
}

@article{du2023instance,
  title={Instance segmentation and berry counting of table grape before thinning based on AS-SwinT},
  author={Du, Wensheng and Liu, Ping},
  journal={Plant Phenomics},
  volume={5},
  pages={0085},
  year={2023},
  publisher={AAAS}
}

@article{gonzalez2025comparison,
  title={Comparison of CNN architectures for single grape detection},
  author={Gonz{\'a}lez, MR and Mart{\'\i}nez-Rosas, ME and Brizuela, CA},
  journal={Computers and Electronics in Agriculture},
  volume={231},
  pages={109930},
  year={2025},
  publisher={Elsevier}
}

@article{palacios2019non,
  title={A non-invasive method based on computer vision for grapevine cluster compactness assessment using a mobile sensing platform under field conditions},
  author={Palacios, Fernando and Diago, Maria P and Tardaguila, Javier},
  journal={Sensors},
  volume={19},
  number={17},
  pages={3799},
  year={2019},
  publisher={MDPI}
}

@article{underhill2020image,
  title={Image-based phenotyping identifies quantitative trait loci for cluster compactness in grape},
  author={Underhill, Anna and Hirsch, Cory and Clark, Matthew},
  journal={Journal of the American Society for Horticultural Science},
  volume={145},
  number={6},
  pages={363--373},
  year={2020},
  publisher={American Society for Horticultural Science}
}

@article{underhill2020image_sys,
author = {A. N. Underhill  and C. D. Hirsch  and M. D. Clark },
title = {Evaluating and Mapping Grape Color Using Image-Based Phenotyping},
journal = {Plant Phenomics},
volume = {2020},
number = {},
pages = {},
year = {2020},
doi = {10.34133/2020/8086309},
URL = {https://spj.science.org/doi/abs/10.34133/2020/8086309}
}

@article{kim2023shine,
  title={A Shine Muscat Grape Berry Detection and Grape Cluster Compactness Estimation for Assessment of Grape Quality Based on Instance Segmentation Methods},
  author={Kim, EungChan and Lee, Chang-Hyup and Park, Seongmin and Hong, Suk-Ju and Kim, Sang-Yeon and Kim, Ghiseok},
  journal={Journal of the ASABE},
  volume={66},
  number={5},
  pages={1175--1185},
  year={2023},
  publisher={American Society of Agricultural and Biological Engineers}
}

@article{seng2018computer,
  title={Computer vision and machine learning for viticulture technology},
  author={Seng, Kah Phooi and Ang, Li-Minn and Schmidtke, Leigh M and Rogiers, Suzy Y},
  journal={IEEE Access},
  volume={6},
  pages={67494--67510},
  year={2018},
  publisher={IEEE}
}

@dataset{morros2021ai4agriculture,
  author       = {Josep Ramon Morros and Tomas Pariente Lobo and Sergio Salmeron-Majadas and Javier Villazan and Diego Merino and Ana Antunes and Mihai Datcu and Chandrabali Karmakar and Edmundo Guerra and Despina-Athanasia Pantazi and George Stamoulis},
  title        = {{AI4Agriculture Grape Dataset} (1.0.0)},
  year         = {2021},
  publisher    = {Zenodo},
  doi          = {10.5281/zenodo.5660081},
  url          = {https://doi.org/10.5281/zenodo.5660081},
  note         = {Data set}
}

@dataset{zabawa2021segmentation,
  author       = {L. Zabawa and A. Kicherer and L. Klingbeil and R. Töpfer and H. Kuhlmann and R. Roscher and M. Pflanz and M. Schirrmann and C. Wellhausen and H. Nordmeyer and R. Töpfer and H. Kuhlmann and R. Roscher},
  title        = {Segmentation of wine berries},
  year         = {2021},
  publisher    = {OpenAgrar Repository},
  volume       = {164},
  doi          = {10.5073/20210308-154150},
  url          = {https://doi.org/10.5073/20210308-154150},
  note         = {Data set}
}

@dataset{isabel_pinheiro_2023_7717055,
  author       = {Isabel Pinheiro},
  title        = {Grapevine Bunch Detection Dataset},
  month        = feb,
  year         = 2023,
  publisher    = {Zenodo},
  doi          = {10.5281/zenodo.7717055},
  url          = {https://doi.org/10.5281/zenodo.7717055},
}

@article{SOZZI2022108466,
title = {wGrapeUNIPD-DL: An open dataset for white grape bunch detection},
journal = {Data in Brief},
volume = {43},
pages = {108466},
year = {2022},
issn = {2352-3409},
doi = {https://doi.org/10.1016/j.dib.2022.108466},
url = {https://www.sciencedirect.com/science/article/pii/S2352340922006606},
author = {Marco Sozzi and Silvia Cantalamessa and Alessia Cogato and Ahmed Kayad and Francesco Marinello}
}

@article{santos2020grape,
  title={Grape detection, segmentation, and tracking using deep neural networks and three-dimensional association},
  author={Santos, Thiago T and De Souza, Leonardo L and dos Santos, Andreza A and Avila, Sandra},
  journal={Computers and Electronics in Agriculture},
  volume={170},
  pages={105247},
  year={2020},
  publisher={Elsevier}
}

@article{BARBOLE2023109100,
title = {GrapesNet: Indian RGB \& RGB-D vineyard image datasets for deep learning applications},
journal = {Data in Brief},
volume = {48},
pages = {109100},
year = {2023},
issn = {2352-3409},
doi = {https://doi.org/10.1016/j.dib.2023.109100},
url = {https://www.sciencedirect.com/science/article/pii/S2352340923002196},
author = {Dhanashree K. Barbole and Parul M. Jadhav}
}

@Article{agronomy13081995,
AUTHOR = {Blekos, Achilleas and Chatzis, Konstantinos and Kotaidou, Martha and Chatzis, Theocharis and Solachidis, Vassilios and Konstantinidis, Dimitrios and Dimitropoulos, Kosmas},
TITLE = {A Grape Dataset for Instance Segmentation and Maturity Estimation},
JOURNAL = {Agronomy},
VOLUME = {13},
YEAR = {2023},
NUMBER = {8},
ARTICLE-NUMBER = {1995},
URL = {https://www.mdpi.com/2073-4395/13/8/1995},
ISSN = {2073-4395},
DOI = {10.3390/agronomy13081995}
}

@Article{jimaging11020034,
AUTHOR = {Quiñones, Rubi and Banu, Syeda Mariah and Gultepe, Eren},
TITLE = {GCNet: A Deep Learning Framework for Enhanced Grape Cluster Segmentation and Yield Estimation Incorporating Occluded Grape Detection with a Correction Factor for Indoor Experimentation},
JOURNAL = {Journal of Imaging},
VOLUME = {11},
YEAR = {2025},
NUMBER = {2},
ARTICLE-NUMBER = {34},
URL = {https://www.mdpi.com/2313-433X/11/2/34},
PubMedID = {39997536},
ISSN = {2313-433X},
DOI = {10.3390/jimaging11020034}
}

@inproceedings{cheng2022masked,
  title={Masked-attention mask transformer for universal image segmentation},
  author={Cheng, Bowen and Misra, Ishan and Schwing, Alexander G and Kirillov, Alexander and Girdhar, Rohit},
  booktitle={Proceedings of the IEEE/CVF conference on computer vision and pattern recognition},
  pages={1290--1299},
  year={2022}
}

@inproceedings{kirillov2023segment,
  title={Segment anything},
  author={Kirillov, Alexander and Mintun, Eric and Ravi, Nikhila and Mao, Hanzi and Rolland, Chloe and Gustafson, Laura and Xiao, Tete and Whitehead, Spencer and Berg, Alexander C and Lo, Wan-Yen and others},
  booktitle={Proceedings of the IEEE/CVF international conference on computer vision},
  pages={4015--4026},
  year={2023}
}

@InProceedings{Wan_2021_CVPR,
    author    = {Wan, Jia and Liu, Ziquan and Chan, Antoni B.},
    title     = {A Generalized Loss Function for Crowd Counting and Localization},
    booktitle = {Proceedings of the IEEE/CVF Conference on Computer Vision and Pattern Recognition (CVPR)},
    year      = {2021},
    pages     = {1974-1983}
}
\end{document}